\documentclass[a4paper]{article}
\usepackage{spconf,amsmath}
\usepackage{amssymb}
\usepackage{subcaption}
\usepackage{tabularx}
\usepackage{url}
\usepackage{multirow}
\usepackage{color}
\usepackage{scrextend}
\graphicspath{ {./images/} }
\usepackage{array}
\newcolumntype{L}[1]{>{\raggedright\let\newline\\\arraybackslash\hspace{0pt}}m{#1}}
\newcolumntype{C}[1]{>{\centering\let\newline\\\arraybackslash\hspace{0pt}}m{#1}}
\newcolumntype{R}[1]{>{\raggedleft\let\newline\\\arraybackslash\hspace{0pt}}m{#1}}

\title{SpeechBERT: An Audio-and-text Jointly Learned Language Model \\for End-to-end Spoken Question Answering}
\name{Yung-Sung Chuang \qquad Chi-Liang Liu \qquad Hung-yi Lee \qquad Lin-shan Lee}
\address{College of Electrical Engineering and Computer Science, National Taiwan University}
\email{\{chuangyungsung,liangtaiwan1230,tlkagkb93901106\}@gmail.com, lslee@gate.sinica.edu.tw}

\begin{document}
\maketitle

\begin{abstract}
While various end-to-end models for spoken language understanding tasks have been explored recently, this paper is probably the first known attempt to challenge the very difficult task of end-to-end spoken question answering (SQA).
Learning from the very successful BERT model for various text processing tasks, here we proposed an audio-and-text jointly learned SpeechBERT model. This model outperformed the conventional approach of cascading ASR with the following text question answering (TQA) model on datasets including ASR errors in answer spans, because the end-to-end model was shown to be able to extract information out of audio data before ASR produced errors. When ensembling the proposed end-to-end model with the cascade architecture, even better performance was achieved. In addition to the potential of end-to-end SQA, the SpeechBERT can also be considered for many other spoken language understanding tasks just as BERT for many text processing tasks. 
\end{abstract}

\vspace{-10pt}
\section{Introduction}
\label{sec:intro}

\vspace{-5pt}

Various spoken language processing tasks, such as translation~\cite{berard2016listen}, retrieval~\cite{lee2015spoken}, summarization~\cite{lu2017order} and understanding~\cite{serdyuk2018towards} have been very successful with a standard cascade architecture: an ASR front-end module transforming the speech signals into text form, followed by the downstream task module (such as translation) trained on text taking the ASR output as normal text input. However, the end-to-end approach trying to consider the two modules as a whole is always attractive for the following reasons. The two modules in the cascade architecture locally optimize the two tasks with different criteria, while the end-to-end approach may obtain globally optimized performance for the overall task. The ASR module minimizes the WER, which is not necessarily directly proportional to the performance measure of the overall task. Much information is inevitably lost when the speech signals are transformed into text with errors, and the errors can't be recovered in the following module. 
The end-to-end approach allows the possibility of capturing information directly from the speech signals not shown in the ASR output and offering overall performance less limited by the ASR accuracy.

Some spoken language tasks such as translation, retrieval, and understanding (intent classification and slot filling) have been achieved with end-to-end approaches~\cite{haghani2018audio, chen2018spoken, palogiannidi2020end, price2020end, huang2020leveraging}, although it remains difficult to obtain significantly better results than the standard cascade architecture. These tasks are primarily sentence-level, in which extracting some local information in a short utterance may be adequate. Spoken question answering (SQA) considered here is known as a much more difficult problem. The inputs to the SQA task are much longer spoken paragraphs. In addition to understanding the literal meaning, the global information in the paragraphs needs to be organized, and sophisticated reasoning is usually required to find the answers. Fine-grained information is also important to predict the exact position of the answer span from a very long context. This paper is probably the first known attempt to try to perform such a difficult SQA task with an end-to-end approach.

Substantial improvements in text question answering (TQA) tasks trained and tested on text data have been observed after the large-scale self-supervised pre-trained language models appeared, such as BERT~\cite{devlin2019bert} and GPT~\cite{Radford2018ImprovingLU}. Instead of learning the TQA tasks from scratch, these models were first pre-trained on a large unannotated text corpus to learn self-supervised representations for the general purpose and then fine-tuned on the downstream TQA dataset. Results comparable to human performance on SQuAD datasets~\cite{squad2016, squad2018} were obtained in this way. However, previous work~\cite{Lee2018spoken} indicated that such an approach may not be easily extendable to the SQA task on audio data by directly cascading an ASR module in the front. Not only the ASR caused serious problems, but very often the true answer spans included name entities or OOV words which cannot be correctly recognized at all, thus cannot be identified by the following TQA module trained on text. That is why all questions with recognition errors in answer spans were discarded in the previous work~\cite{Lee2018spoken}. This led to the end-to-end approach to SQA proposed here.

The BERT model~\cite{devlin2019bert} useful in TQA tasks was able to transform word tokens into contextualized embeddings carrying plenty of information. For SQA task, it is certainly desirable to transform audio words (audio signals for word tokens) also into such embeddings, but much more challenging. The same word token can have millions of different audio signal realizations in different utterances. 
The boundaries for audio words in utterances are not available. 
The next problem is even much more challenging. BERT can learn semantic information of a word token based on its context in text form. Audio words in audio data have context only in audio form, which is much more noisy, confusing, unpredictable, and difficult to handle. Learning semantics from audio context is really hard.

Audio Word2Vec~\cite{chung2016audio} was the first effort to transform audio words with known boundaries into embeddings carrying phonetic information only, no semantics at all. Speech2Vec~\cite{speech2vec2018} then tried to imitate the training process of skip-gram or CBOW in Word2Vec~\cite{mikolov2013distributed} to extract some semantic features. It was proposed~\cite{wang2018segmental} to learn to jointly segment the audio words out of utterance and extract the embeddings. Other efforts then tried to align the audio word embeddings with text word embeddings~\cite{yichen2018, chung2019nips}. Some more recent works~\cite{liu2019mockingjay, baevski2019vq, jiang2019improving, song2019speech, fan2019unsupervised, ling2019deep} even tried to obtain embeddings for audio signals using approaches very similar to BERT, but primarily extracting acoustic information with very limited semantics. These approaches may be able to extract some semantics with audio word embeddings, but the level of semantics obtained was still very far from that required for the SQA tasks.

In this paper, we propose an audio-and-text jointly learned \textbf{SpeechBERT} model for the end-to-end SQA task. SpeechBERT is a pre-trained model learned from audio and text datasets, so as to be able to produce embeddings for audio words, and these embeddings can be properly aligned with the embeddings for the corresponding text words offered by a typical text BERT model trained with text datasets. Standard pre-training and fine-tuning as text BERT are also performed. When used in the SQA task, performance comparable to or better than the cascade architecture was obtained.

\vspace{-5pt}
\section{SpeechBERT for End-to-end SQA}
\label{sec:SpeechBERT}

\begin{figure}[t!]
    \centering
    \includegraphics[width=0.9\linewidth]{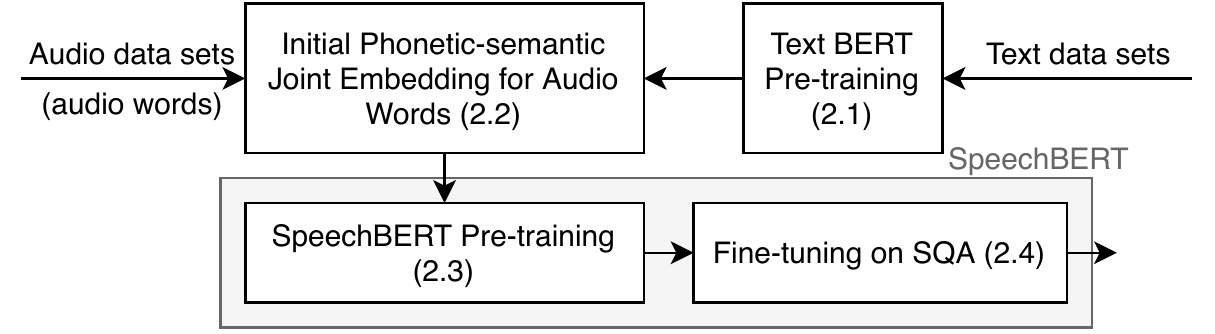}
    \caption{Overall training process for SpeechBERT.}
    \label{fig:1}
    \vspace{-15pt}
\end{figure}
\vspace{-5pt}

Here we assume the training audio datasets include the ground truth transcripts, and an off-the-shelf ASR engine is available. So we can segment the audio training data into audio words (audio signals for the underlying word tokens) by forced alignment. For audio testing data we can use the ASR engine to obtain audio words including their boundaries too, although with errors. The overall training process for the SpeechBERT for end-to-end SQA proposed here is shown in Fig.~\ref{fig:1}. We first use the text dataset to pre-train a Text BERT (upper right block, Sec.~\ref{ssec:bert}), based on which we train the initial phonetic-semantic joint embedding with the audio words in the training audio dataset (upper left block, Sec.~\ref{ssec:sse}). The training of SpeechBERT (or a shared BERT model for both text and audio) is then in the bottom of the figure, including pre-training (Sec~\ref{ssec:trainproc}) and fine-tuning on SQA (Sec.~\ref{ssec:finetune}). The details are given below.

\vspace{-6pt}
\subsection{Text BERT Pre-training}
\label{ssec:bert}
\vspace{-3pt}

Here we follow the standard Text BERT pre-training procedure~\cite{devlin2019bert}. For a sentence in the training text set with $n$ tokens $T = \{t_1, t_2, ..., t_n\}$, we represent them as vectors $E_{\text{text}} = \{e_1, e_2, ..., e_n\}$ and sum them with corresponding positional and sentence segment embeddings to get $E_{\text{text}}^\prime$ to be fed into the multi-layer Transformer. Masked language model (MLM) task is performed at the output layer of the Text BERT by randomly replacing 15\% of vectors in $E_{\text{text}}$ with a special mask token vector and predicting the masked tokens at the same position of the output layers. Next sentence prediction (NSP) usually performed in BERT training is not used here because some recent studies indicated it may not be helpful~\cite{xlm2019, spanbert2019, xlnet2019, roberta2019}.

\vspace{-8pt}
\subsection{Initial Phonetic-Semantic Joint Embedding}
\label{ssec:sse}
\begin{figure}[t!]
    \centering
    \includegraphics[width=0.9\linewidth]{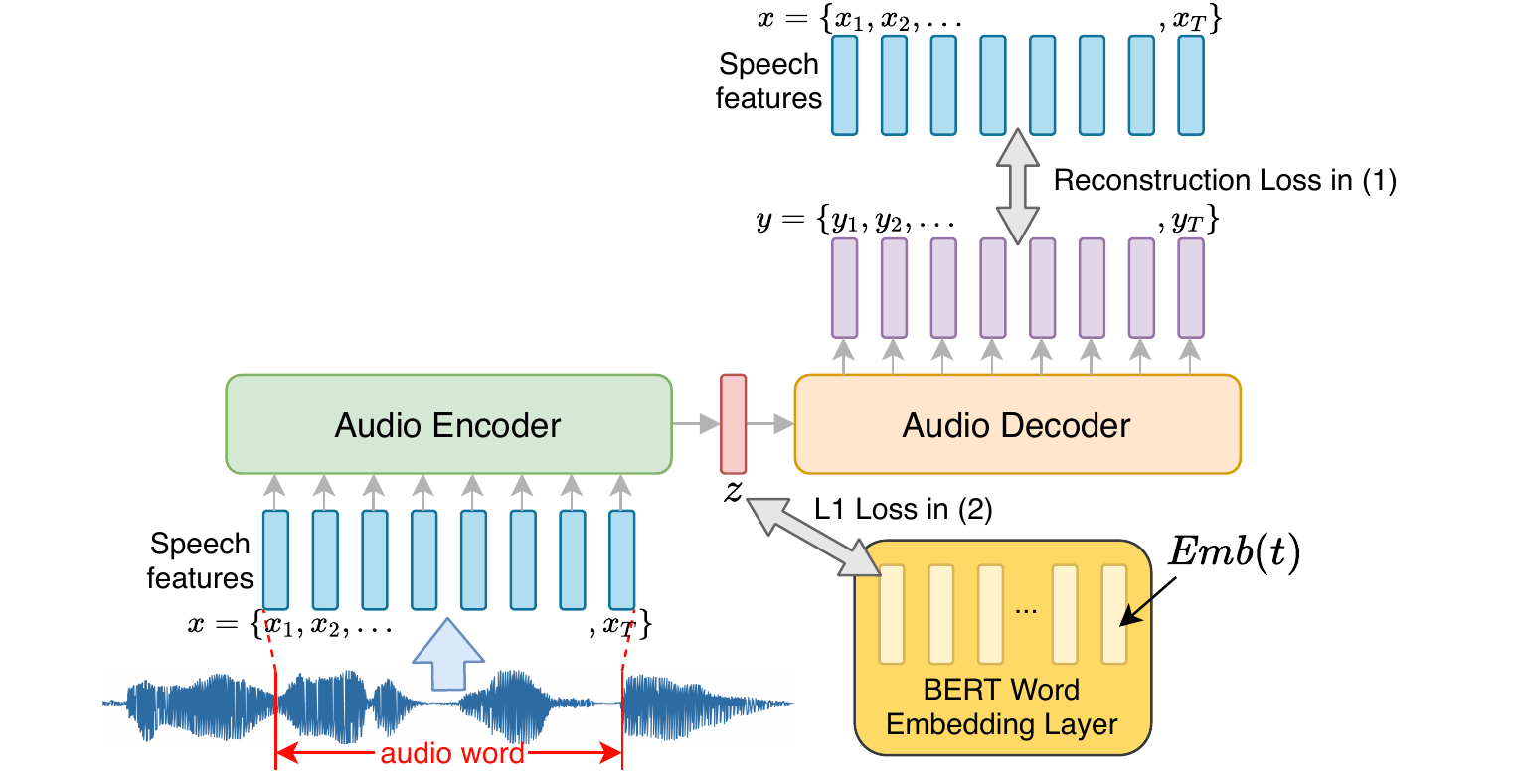}
    \caption{Training procedure for the Initial Phonetic-Semantic Joint Embedding. After training, the the encoded vector ($z$ in red) obtained here is used to train the SpeechBERT.}
    \label{fig:audio2vec}
    \vspace{-15pt}
\end{figure}

\vspace{-3pt}
For an utterance in the training audio dataset with $n$ audio words $X = \{x^{(1)}, x^{(2)}, ..., x^{(n)}\}$, the goal here is to encode these $n$ audio words into $n$ embeddings $E_{\text{audio}} = \{\tilde{e_1}, \tilde{e_2}, ..., \tilde{e_n}\}$. Let one of the audio words include a total of $T$ speech feature vectors, $x = \{x_1, x_2, ..., x_T\}$. The complete process of initial phonetic-semantic joint embedding is in Fig.~\ref{fig:audio2vec}. 
With the speech features for each audio word $x = \{x_1, x_2, ..., x_T\}$, we use an RNN sequence-to-sequence autoencoder~\cite{chung2016audio} as in Fig.~\ref{fig:audio2vec}. This includes an audio encoder (low left corner of the figure) transforming the input $x$ into a vector $z$ (in red in the middle), and an audio decoder reconstructing the output $y = (y_1, y_2, ..., y_T)$ from $z$. The autoencoder is trained to minimize the reconstruction error:
\vspace{-10pt}
\begin{equation}
\mathcal{L}_{\rm recons} = \sum_{k}\sum_{t=1}^{T}\left\|x_{t}-y_{t}\right\|^{2}_2,
\vspace{-5pt}
\end{equation}
where $k$ is the index for training audio words. This process enables the vector $z$ to capture the phonetic structure information of the audio words but not semantics at all, which is not adequate for the goal here. So we make the vector $z$ be constrained by a L1-distance loss (lower right) further:
\vspace{-5pt}
\begin{equation}
\mathcal{L}_{L_1} = \sum_{k}\left\|z-\textit{Emb}(t)\right\|_{1},
\vspace{-7pt}
\end{equation}
where $t$ is the token label for the audio word $x$ and $\textit{Emb}$ is the embedding layer of the Text BERT trained in Sec.~\ref{ssec:bert}. In this way, we use the token label of the audio word to access the semantic information about the considered audio word extracted by the Text BERT as long as it is within the vocabulary of the Text BERT. So, the autoencoder model learns to keep the phonetic structure of the audio word so as to reconstruct the original speech features $x$ as much as possible, but at the same time it learns to fit to the embedding distribution of the Text BERT which carries plenty of semantic information for the word tokens. This makes it possible for the model to learn a joint embedding space integrating both phonetic and semantic information extracted from both audio and text datasets.

\vspace{-5pt}
\subsection{MLM Pre-training for SpeechBERT with Both Text and Audio Data}
\label{ssec:trainproc}

\begin{figure*}[t!]
    \centering
    \begin{subfigure}{\columnwidth}
        \centering
        \includegraphics[width=1.0\linewidth]{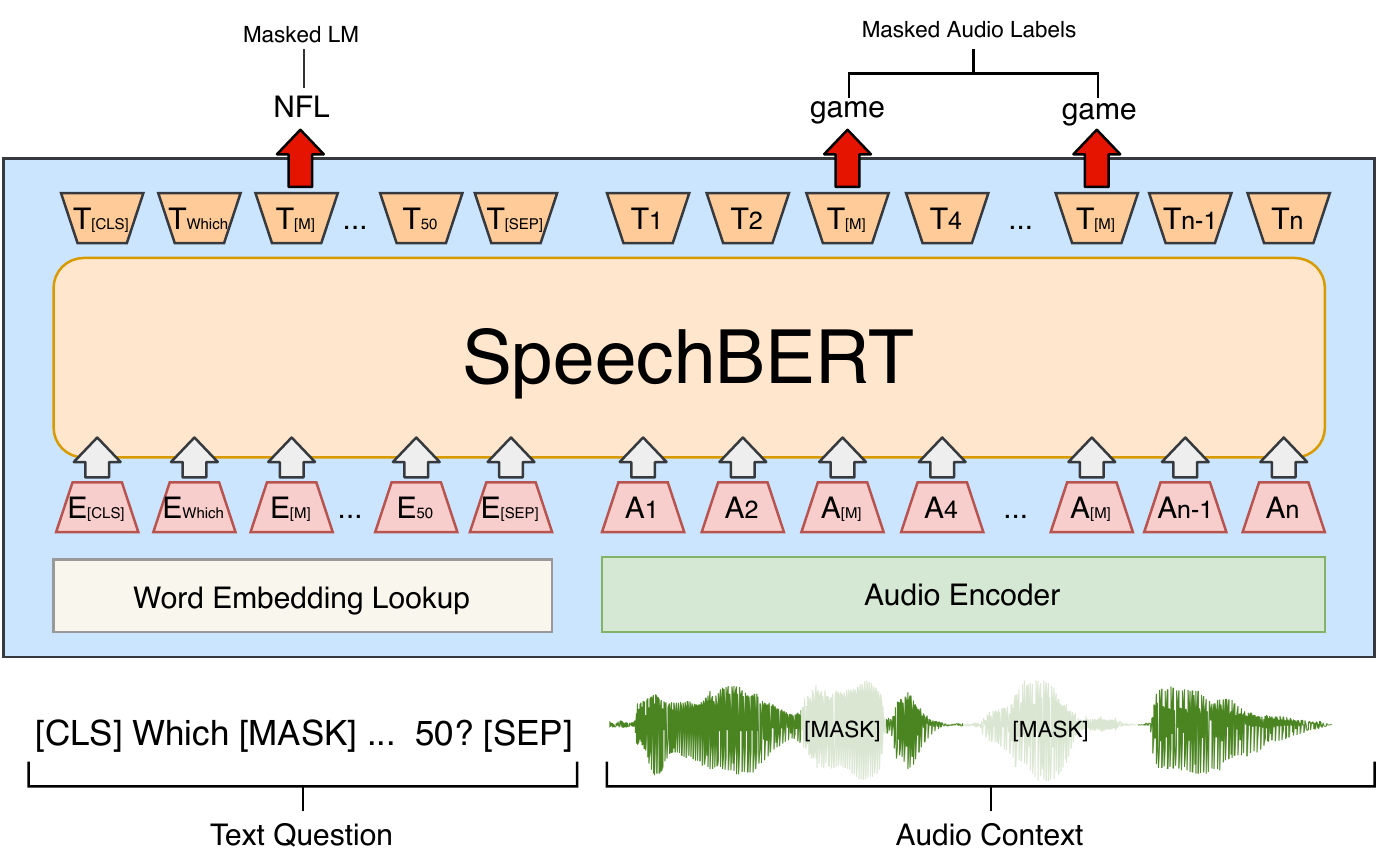}   
        \caption{Pre-training Stage}
    \end{subfigure}
    \begin{subfigure}{\columnwidth}
        \centering
        \includegraphics[width=1.0\linewidth]{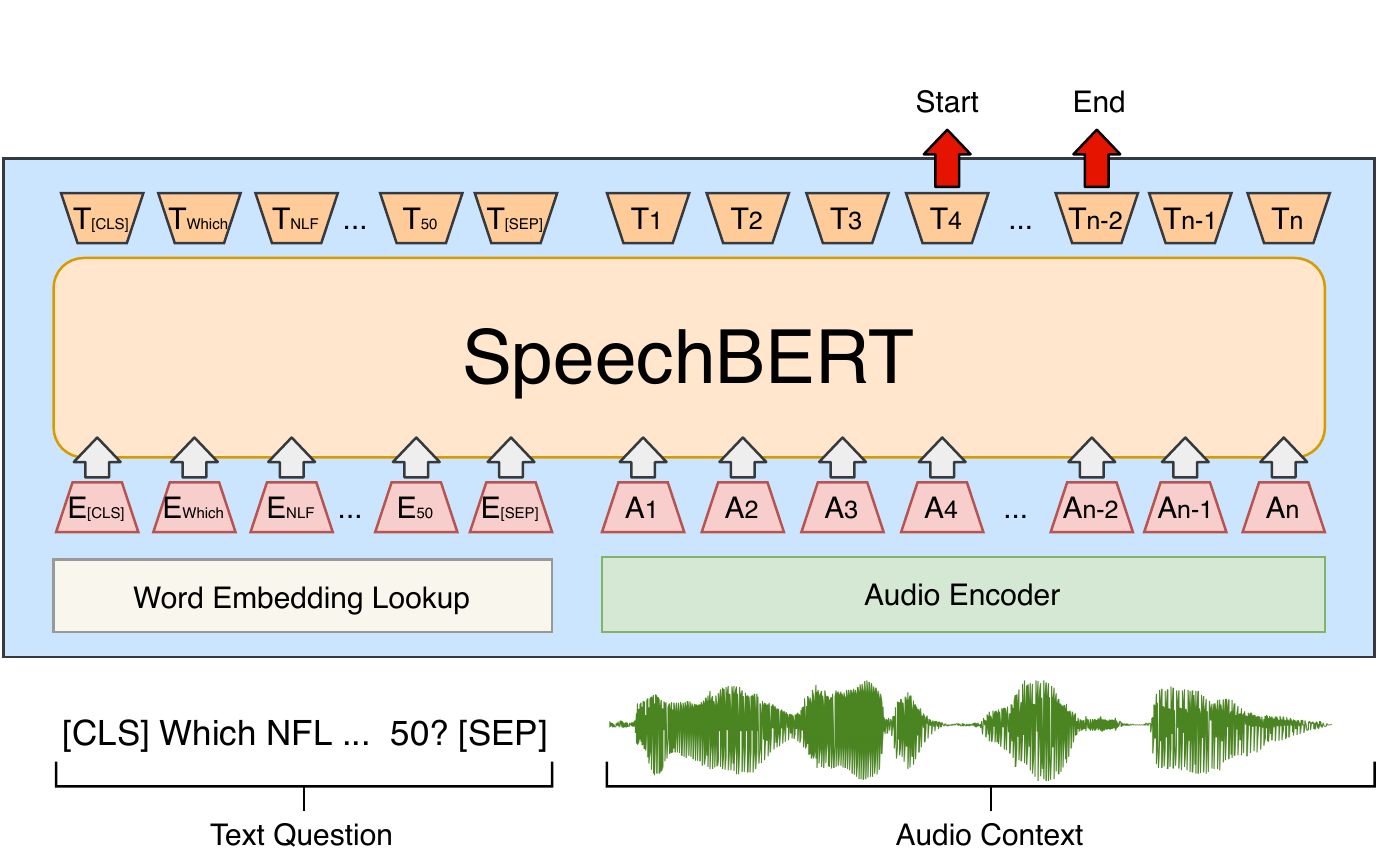}
        \caption{Fine-tuning Stage}
    \end{subfigure}
    \caption{Two training stages for the SpeechBERT model: (a) Pre-training and (b) Fine-tuning. The two stages use identical model architecture except for the output layers. The special tokens [CLS] and [SEP] are added following the original Text BERT.}
    \label{fig:speechbert}
    \vspace{-15pt}
\end{figure*}

\vspace{-5pt}
This is the lower left block of Fig.~\ref{fig:1} with details in Fig.~\ref{fig:speechbert} (a). Here we pre-train the SpeechBERT with the MLM task using both text and audio datasets, before fine-tuning it on the downstream QA task.

As in Fig.~\ref{fig:speechbert} (a), for the SpeechBERT learning to take embeddings for both discrete text words and continuous spoken words, we jointly optimize the MLM loss with the mixture of both audio and text data. The same training target for text MLM as described in Sec.~\ref{ssec:bert} is used here. For audio data input, after obtaining the phonetic-semantic joint embeddings $E_{\text{audio}} = \{\tilde{e_1}, \tilde{e_2}, ..., \tilde{e_n}\}$ for each utterance as in Sec.~\ref{ssec:sse}, we also randomly replace 15\% of those embeddings with mask token vectors as we do in Sec.~\ref{ssec:bert}. With the supervised setting, we can similarly predict the corresponding tokens behind the masked spoken words. During training, we freeze the audio encoder in Fig.~\ref{fig:audio2vec} to speed up the process, while have the text word embedding unfrozen to keep the joint audio-and-text embeddings flexible, considering the earlier experiences reported for end-to-end SLU~\cite{Lugosch2019}.

\vspace{-5pt}
\subsection{Fine-tuning on Question Answering}
\label{ssec:finetune}

\vspace{-5pt}
This is the last block in Fig.~\ref{fig:1}, with details in Fig.~\ref{fig:speechbert} (b). Here the downstream QA task is fine-tuned to minimize the loss for predicting the correct start/end positions of the answer span, as proposed in BERT~\cite{devlin2019bert}. By introducing a start and an end vector $S, E \in \mathbb{R}^{H}$, we compute a dot product for $S$ or $E$ with each final hidden vector $T_i \in \mathbb{R}^{H}$  for audio word $i$ from the SpeechBERT, as in Fig.~\ref{fig:speechbert} (b). The dot product value is softmax-normalized over all audio words in the utterance to compute the probability of audio word $i$ being the start or end position.

\vspace{-5pt}
\section{Experimental Setup}
\label{sec:expsetup}

\vspace{-5pt}
\subsection{Data and Evaluation}
\label{ssec:data}

\begin{figure}[t!]
    \centering
    \includegraphics[width=\linewidth]{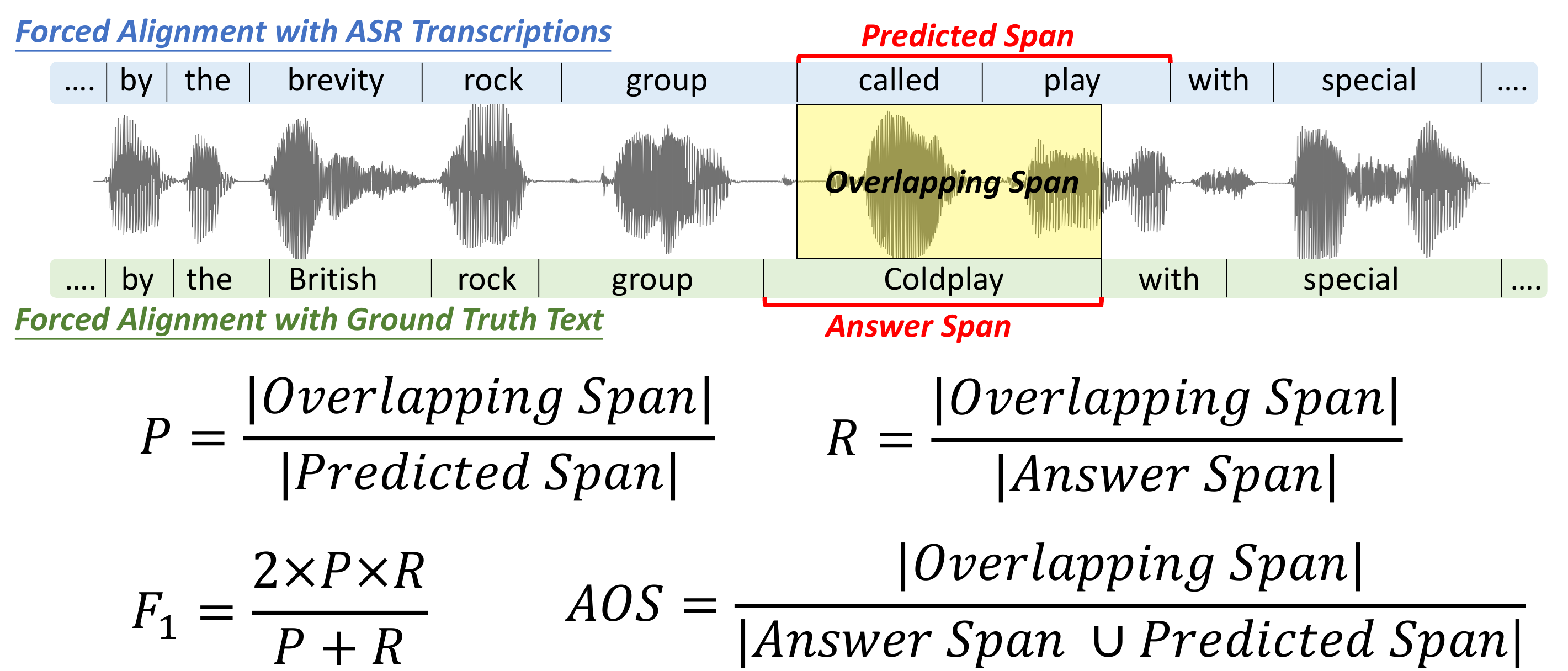}
    \caption{Illustration of the way to evaluate frame-level F1-score and AOS on SQuAD-lost. If the predicted span overlaps well with the ground truth answer span, high F1 and AOS will be obtained even with recognition errors.}
    \label{fig:spokenalign}
\vspace{-20pt}
\end{figure}

\vspace{-5pt}
The SpeechBERT model was trained on the Spoken SQuAD dataset~\cite{Lee2018spoken}, in which the text for all audio paragraphs and all questions are from original SQuAD~\cite{squad2016} dataset. It also includes SQuAD format ASR transcripts plus 37k question-answer pairs as the training set and 5.4k as the testing set. This is smaller than the official SQuAD dataset (with 10.6k questions in its development set) since 5.2k of questions in SQuAD for which the answers couldn't be found in the ASR transcripts were removed and not shown in the Spoken SQuAD testing set. These removed questions in SQuAD were collected to form another testing set referred to as \textbf{SQuAD-lost} (lost by ASR errors). This is helpful below in evaluating whether the end-to-end approach can better handle the questions with incorrectly recognized answer spans. All questions in SQuAD-lost share the same audio files as the Spoken SQuAD testing set.

The evaluation metrics we used are the Exact Matched (EM) percentage and F1 score for the word tokens as in the normal QA tasks, but for SQuAD-lost frame-level F1 and Audio Overlapping Score (AOS)~\cite{Lee2018spoken} based on the boundaries of the answer spans as illustrated in Figure~\ref{fig:spokenalign} were used instead. For the latter case, the boundaries for the audio words were found by forced alignment with Kaldi~\cite{Povey2011}, based on which the start/end points of the answer for both the ground truth and predicted results in training and testing sets were obtained. The ground truth text of the testing set was used only in evaluation.

\vspace{-10pt}
\subsection{Model Setting and Training}
\label{ssec:model}
\vspace{-5pt}

\subsubsection{Initial Phonetic-semantic Joint Embedding}
\label{ssec:expsse}
\vspace{-5pt}

For the autoencoder in Fig.~\ref{fig:audio2vec}, we used a bidirectional LSTM as the encoder and a single-directional LSTM as the decoder, both with input size 39 (MFCC-dim) and hidden size 768 (BERT embedding-dim). Two layers of the fully-connected network are added at the encoder output to transform the encoded vectors to fit the BERT embedding space. We directly used the audio from Spoken SQuAD training set to train this autoencoder.

\vspace{-5pt}
\subsubsection{Text BERT and SpeechBERT}
\label{ssec:expbert}
\vspace{-5pt}

We used the PyTorch implementation\footnote{\label{pytorch}\url{https://github.com/huggingface/transformers}} of BERT to build the BERT model with 12 layers of \texttt{bert-base-uncased} setting. We randomly initialized a new embedding layer with our vocabulary set counted in the dataset rather than using the WordPiece tokenizer~\cite{wu2016} in processing the text because it is inconsistent with the audio word units. The official pre-trained weights were loaded as the weights of the BERT model. We first trained the MLM task with the text of Spoken SQuAD training set for three epochs. 
Next, we used the Spoken SQuAD training set by directly feeding text and audio into the BERT model. 
After pre-training, we fine-tuned the model with the Spoken SQuAD training set for another two epochs. The other hyper-parameters used were identical to the original PyTorch implementation\footref{pytorch}.

\vspace{-5pt}
\section{Experimental Results}
\label{sec:expresults}

\makeatletter
\newcommand{\hwidth}[1]{%
  \noalign{\hrule \@height #1}%
}
\makeatother

\begin{table}[]
\centering
\caption{Experimental results on Spoken SQuAD. Sections (I)(II)(III) are respectively for models trained on text, end-to-end models trained on audio, and ensembled models; while column (A)(B) are respectively for text and ASR testing sets.}
\vspace{-5pt}
\label{tab:stateoftheart}
\setlength\tabcolsep{4.5pt}
\renewcommand{\arraystretch}{1.25}
\begin{tabular}{|ccc|cc|}

\hwidth{1pt}
\multicolumn{1}{|c||}{\multirow{2}{*}{\textbf{Models and Training set}}} & \multicolumn{4}{c|}{\textbf{Testing Set}}  \\
 \cline{2-3}\cline{4-5}

\multicolumn{1}{|c||}{\multirow{2}{*}{}} &
\multicolumn{2}{c|}{\textbf{(A) Text}} & \multicolumn{2}{c|}{\textbf{(B) ASR}}  \\

\hwidth{.4pt}
\multicolumn{1}{|c||}{\textbf{ (I) trained on text}} & EM & F1 & EM & F1 \\

\hwidth{.4pt}

\multicolumn{1}{|l||}{(a) BiDAF on Text~\cite{seo2016bidirectional}} & 58.40 & 69.90 & 37.02 & 50.90   \\

\multicolumn{1}{|l||}{(b) Dr.QA on Text~\cite{chen2017reading}} & 62.84 & 73.74 & 41.16 & 54.51\\

\multicolumn{1}{|l||}{(c) BERT on Text~\cite{devlin2019bert}} & \textbf{76.90} & \textbf{85.71} & 53.30 & 66.17 \\

\multicolumn{1}{|l||}{(d) BERT on ASR~\cite{devlin2019bert}\scriptsize{(cascade)}} & - & - & \textbf{56.28} & \textbf{68.22} \\

\hwidth{.4pt}

\multicolumn{3}{|c||}{\textbf{(II) End-to-end trained on Audio}} & EM & F1 \\

\hwidth{.4pt}
\multicolumn{3}{|l||}{(e) SpeechBERT (proposed)} & \textbf{51.19} & \textbf{64.08} \\

\multicolumn{3}{|l||}{(f) SpeechBERT \textit{w/o MLM}} & 46.02 & 59.62 \\
\multicolumn{3}{|l||}{(g) SpeechBERT \textit{tested on better boundaries}} & \textbf{53.42} & \textbf{66.27} \\

\hwidth{.4pt}

\multicolumn{3}{|c||}{\textbf{(III) Ensembled models}} & EM & F1 \\

\hwidth{.4pt}
\multicolumn{3}{|l||}{(h) ensembled [(e) plus (d)]} & \textbf{60.37} & \textbf{71.75} \\
\multicolumn{3}{|l||}{(i) ensembled [(d) plus (d)]} & 57.88 & 69.29 \\

\hwidth{1pt}
\end{tabular}
\end{table}

\begin{table}[]
\centering
\caption{Experimental results on SQuAD-lost, Spoken SQuAD and Total for the cascade and end-to-end models.}
\vspace{-5pt}
\label{tab:hidden}

\setlength\tabcolsep{3pt}
\renewcommand{\arraystretch}{1.25}
\begin{tabular}{|c|cc|cc|cc|}

\hwidth{1pt}

\multirow{2}{*}{\textbf{Model}} & \multicolumn{2}{c|}{\textbf{Spoken SQuAD}}&
\multicolumn{2}{c|}{\textbf{SQuAD-lost}} &  \multicolumn{2}{c|}{\textbf{Total}} \\
\cline{2-3}\cline{4-5}\cline{6-7}

\multirow{2}{*}{} & \textbf{F1} & \textbf{AOS} & \textbf{F1} & \textbf{AOS} & \textbf{F1} & \textbf{AOS} \\

\hwidth{0.4pt}
Cascade (d) & \textbf{66.56} & \textbf{63.87} & 30.78 & 27.58 &  48.74 & 45.84\\
End-to-end (e) & 62.76 & 59.70 & \textbf{37.31} & \textbf{33.57} & \textbf{50.12} & \textbf{46.72}\\

\hwidth{1pt}
\end{tabular}
\vspace{-15pt}
\end{table}

\subsection{Spoken SQuAD (no ASR errors in answer spans)}
\vspace{-5pt}

In Table~\ref{tab:stateoftheart} Section (I) is for models trained on text data, with rows (a)(b)(c) for different QA models trained on the same SQuAD (in text), while column (A)(B) respectively for testing sets of SQuAD (text) and Spoken SQuAD (ASR). Rows (a)(b)(c) showed the prior arts and the superiority of BERT (rows (c) vs (a)(b)), and the serious performance degradation for these models when tested on ASR output directly (column (B) vs (A)). Row (d) is for BERT trained on ASR transcriptions of Spoken SQuAD, or the ``cascade'' of ASR and BERT, where we see BERT performed much better for ASR output if trained on ASR output (rows (d) vs (c) for column (B)).

Section (II) is for the end-to-end QA model proposed here, with row (e) for the SpeechBERT trained on Spoken SQuAD. We see the end-to-end model is still 4-5\% lower than the ``cascade'' architecture (row (e) vs (d)), although much better or comparable to prior models directly used on ASR output (rows (e) vs (a)(b)(c)). Considering the high level of the difficulty for the end-to-end SQA model to learn the sophisticated semantic knowledge out of the relatively long audio signals directly in one model without using the word tokens from ASR, the results here showed the promising potential for end-to-end approaches for SQA when better data, better model and better training become possible.

When we further ensembled the end-to-end model in row (e) with the cascade architecture in row (d) as shown in row (h) of Section (III), we see significantly improved performance compared to the two component models (rows (h) vs (d) or (e)), achieving the state-of-the-art result on Spoken SQuAD. Note that we can also ensemble two separately trained cascade models [(d) plus (d)] as shown in row (i), but the achievable improvement was much less (row (i) vs (h)). These results showed that the end-to-end SpeechBERT can learn extra knowledge complementary to that learned by the cascade architecture. Furthermore, the results in row (h) and column (B) trained and tested on audio are already higher than those  in row (a) and column (A) on ground truth text and comparable to those in row (b) and column (A), although still much lower than row (c) and column (A). This implies that the results obtained here are substantial and significant.

\vspace{-5pt}
\subsection{SQuAD-lost (with ASR errors in answer spans)} 
\label{ssec:ensemble}
\vspace{-5pt}

We wish to find out further with the end-to-end model if we can better handle the questions with incorrectly recognized answer spans. Here we used the frame-level F1 and AOS as illustrated in Fig.~\ref{fig:spokenalign} for evaluation and the results for cascade (row (d) in Table~\ref{tab:stateoftheart}) and end-to-end (row (e) in Table~\ref{tab:stateoftheart}) are in Table~\ref{tab:hidden},  SQuAD-lost  (incorrectly recognized answer spans), Spoken SQuAD (same as in Table~\ref{tab:stateoftheart}, all with correctly recognized answer spans), and the total of the two. The results show the end-to-end models did significantly better on SQuAD-lost (middle), although worse by a gap on Spoken SQuAD (left), but offered some improvements on the total (right). This verified the end-to-end model could learn some phonetic-semantic knowledge directly from audio signals before errors occurred in ASR, and explained indirectly why the ensembled model ([(e) plus (d)] in row (h)) of Table~\ref{tab:stateoftheart} can do better than cascade alone (row (d)). The scenario on the total on the right of Table~\ref{tab:hidden} was closer to the real-world applications.

\vspace{-5pt}
\subsection{Further Analysis} 
\vspace{-5pt}
\label{ssec:abs}

\subsubsection{Ablation MLM pre-training and Better Word Boundaries}
\vspace{-5pt}

The results in row (f) of Section (II) in Table~\ref{tab:stateoftheart} are for the end-to-end model trained directly with SQA fine-tuning without the MLM pre-training, and significant performance drop can be observed (rows (f) vs (e)). Also, the results in row (e) of Table~\ref{tab:stateoftheart} are for audio word boundaries obtained by forced alignment using ASR transcripts with WER of 22.73\% in Spoken SQuAD~\cite{Lee2018spoken}. We further tested the model on \emph{better boundaries} obtained with forced alignment using 
\emph{ground truth text}. By doing so, we can see the extent to which performance is limited by segmentation quality. The results in row (g) showed some improvements could still be achieved with better boundaries (rows (g) vs (e)), which can be a direction for future work.

\vspace{-10pt}
\subsubsection{Analysis on WER of ASR}
\label{ssec:ea}

\begin{figure}[t!]
    \centering
    \includegraphics[width=0.9\linewidth]{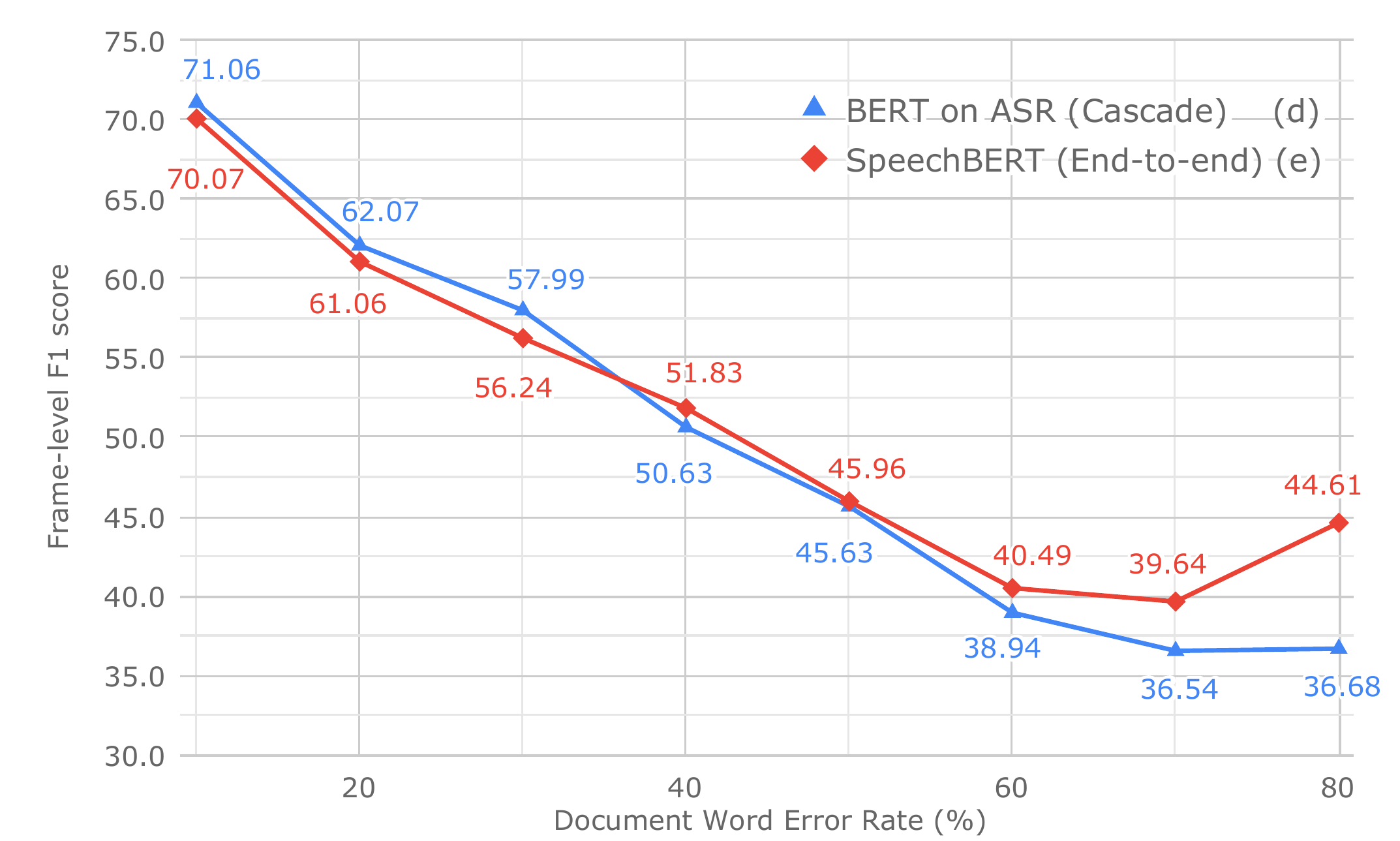}
    \caption{Frame-level F1 scores evaluated for small groups of Total (Spoken SQuAD/SQuAD-lost) at different levels of WER.}
    \label{fig:emratio}
    \vspace{-15pt}
\end{figure}

\vspace{-5pt}
To investigate how the cascade and end-to-end models (rows (d) and (e) in Table~\ref{tab:stateoftheart}) worked with audio at different WER, we split the questions in the total dataset including both SQuAD-lost and Spoken SQuAD into smaller groups with different WER. The frame-level F1 results for these groups were plotted in Figure~\ref{fig:emratio}. Obviously, at lower WER both models offered higher F1, and the cascade architecture performed better. Both models suffered from performance degradation at higher WER, and the end-to-end model outperformed cascade when WER exceeded 40\% since it never relied on ASR output.

\vspace{-5pt}
\section{Concluding Remarks}
\label{sec:conclusions}

\vspace{-5pt}

Audio signals are in form of phonetic structures, while carrying semantics. ASR has long been used to transform the phonetic structures into word-level tokens carrying semantics, but with inevitable errors causing troubles to the downstream application tasks. With reasonable performance of the end-to-end SQA task, the proposed SpeechBERT was shown to be able to do similar transformation, but the semantics were somehow directly tuned to the downstream task, or question answering here, bypassing the intermediate problem of ASR errors. This concept (and the SpeechBERT) is definitely useful to many other spoken language processing tasks to be studied in the future.

\section{Acknowledgements}

We are grateful to the National Center for High-performance Computing for computer time and facilities.

\bibliographystyle{IEEEbib}
\bibliography{refs}

\end{document}